\documentclass{llncs}

\usepackage[utf8]{inputenc}
\usepackage{graphicx}
\usepackage{float}
\usepackage[super]{nth}
\usepackage{multirow}
\usepackage{comment}
\date{May 2020}

\begin{document}

\title{A Metamodel and Framework For AGI}
\author{Hugo Latapie\inst{1} \and
Ozkan Kilic\inst{2}}

\begin{comment}
\authorrunning{H. Latapie and O. Kilic}
\end{comment}

%
\institute{Cisco Systems\\
\email{hlatapie@cisco.com}\\
\and
Cisco Systems\\
\email{okilic@cisco.com}}

\maketitle
\begin{abstract}
Can artificial intelligence systems exhibit superhuman performance, but in critical ways, lack the intelligence of even a single-celled organism? The answer is clearly `yes' for narrow AI systems. Animals, plants, and even single-celled organisms learn to reliably avoid danger and move towards food. This is accomplished via a physical knowledge-preserving metamodel that autonomously generates useful models of the world. We posit that preserving the structure of knowledge is critical for higher intelligences that manage increasingly higher levels of abstraction, be they human or artificial. This is the key lesson learned from applying AGI subsystems to complex real-world problems that require continuous learning and adaptation. In this paper, we introduce the Deep Fusion Reasoning Engine (DFRE), which implements a knowledge-preserving metamodel and framework for constructing applied AGI systems. The DFRE metamodel exhibits some important fundamental knowledge preserving properties such as clear distinctions between symmetric and anti-symmetric relations, and the ability to create a hierarchical knowledge representation that clearly delineates between levels of abstraction. The DFRE metamodel, which incorporates these capabilities, demonstrates how this approach benefits AGI in specific ways such as managing combinatorial explosion and enabling cumulative, distributed and federated learning. Our experiments show that the proposed framework achieves 94\% accuracy on average on unsupervised object detection and recognition. This work is inspired by the state-of-the-art approaches to AGI, recent AGI-aspiring work, the granular computing community, as well as Alfred Korzybski's general semantics.

\end{abstract}

\section{Introduction}
After over 30 years of intense effort, the AGI community has developed the theoretical underpinnings for AGI and affiliated working software systems \cite{wang2005,wang2006}. While achieving  human-level AGI is  arguably years to decades away, some of the currently available AGI subsystems are ready to be incorporated into non-profit and for-profit products and services. Some of the most promising AGI systems we have encountered are OpenNARS \cite{wang2006,wang2010}, OpenCog \cite{goertzel2009,goertzel2013}, and AERA \cite{thorisson2012}. We are collaborating with all three teams and have developed video analytics and smart city applications that leverage both OpenCog  and OpenNARS \cite{hammer2019}.

After several years of applying these AGI technologies to complex real-world problems in IoT, networking, and security at scale, we have encountered a few stumbling blocks largely related to real-time performance on large datasets and cumulative learning \cite{thorisson2019}. In order to move from successful proofs of concept and demos to scalable products, we have developed the DFRE metamodel and associated DFRE framework, which is the focus of this paper. We have used this metamodel and framework to bring together a wide array of technologies ranging from machine learning, deep learning, and probabilistic programming to the reasoning engines operating under the Assumption of Insufficient Knowledge and Resources (AIKR) \cite{wang2005}. As discussed below, we believe the initial results are promising: Data show  a dramatic increase in system accuracy, ability to generalize, resource utilization, and real-time performance.

\begin{figure}[h]
\includegraphics[width=\textwidth]{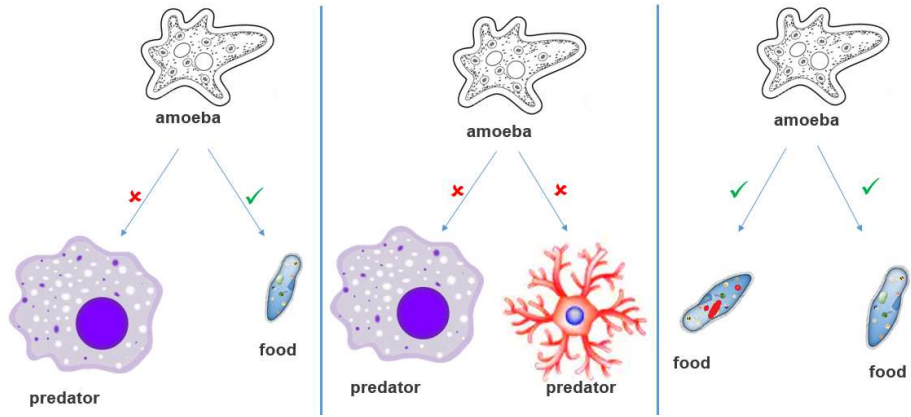}
\caption{\label{fig:amoeba}Amoeba distinguishing between distinctions and similarities.}
\end{figure}

The DFRE metamodel and framework are based on the idea that knowledge is structure \textemdash More specifically, that knowledge is hierarchical structure, where there are levels corresponding to levels of abstraction. The DFRE metamodel refers to how knowledge is hierarchically structured while a model refers to knowledge stored in a manner complying to the DFRE metamodel. It is based on non Aristotelian, non-elementalistic, system thinking. The backbone of its hierarchical structure is based on difference, aka antisymmetric relations, while the leaves of this structure are based on symmetric relations. As in figure \ref{fig:amoeba}, even a simple amoeba has to keep track of distinctions and similarities because not preserving symmetric and antisymmetric relations is fatally important.

Korzybski \cite{korzybski1994}  dedicated the majority of his professional life to analyzing and studying the nature of this hierarchical structure. While it is well beyond the scope of this paper to discuss the details of his analysis, our initial focus was to incorporate these fundamental principles.
\begin{itemize}
  \item K1 – the core framework of knowledge is based on anti-symmetric relations
        \begin{itemize}
            \item Spatial understanding: right/left/top/bottom
            \item Temporal understanding: before/after
            \item Corporal understanding: pain/satiation
            \item Emotional understanding: happy/sad
            \item Social understanding: friend/foe
            \item Causal understanding: X causes Y

        \end{itemize}
  \item K2 – symmetric relations add further structure
        \begin{itemize}
            \item A and B are friends
            \item A is like B
        \end{itemize}
  \item K3 – knowledge is layered
        \begin{itemize}
            \item Sensor data is on a different layer than high level symbolic information
            \item Symbolic information B, which expands or provides context to symbolic information A, is at a higher layer / level of abstraction
            \item In the symbolic space, there are theoretically an infinite number of layers i.e. it is always possible to refer to a symbol and expand upon it, thus creating yet another level of abstraction
        \end{itemize}
  \item K4 – since knowledge is structure, any structure destroying operations such as confusing levels of abstraction, treating an anti-symmetric relation as symmetric, or vice-versa, can be classified as a knowledge corrupting and/or destroying operation \footnote[1]{In [general semantics] Korzybski argues that what is currently limiting humanity’s advancement is the general lack of understanding of how our own abstracting mechanisms work. He considered mankind to currently be in the childhood of humanity and the day, if it should come, that humanity becomes generally aware of the metamodel, is the day humanity enters into the “manhood of humanity” \cite{korzybski1921}} . 
\end{itemize}

\begin{figure}[h]
\includegraphics[width=\textwidth]{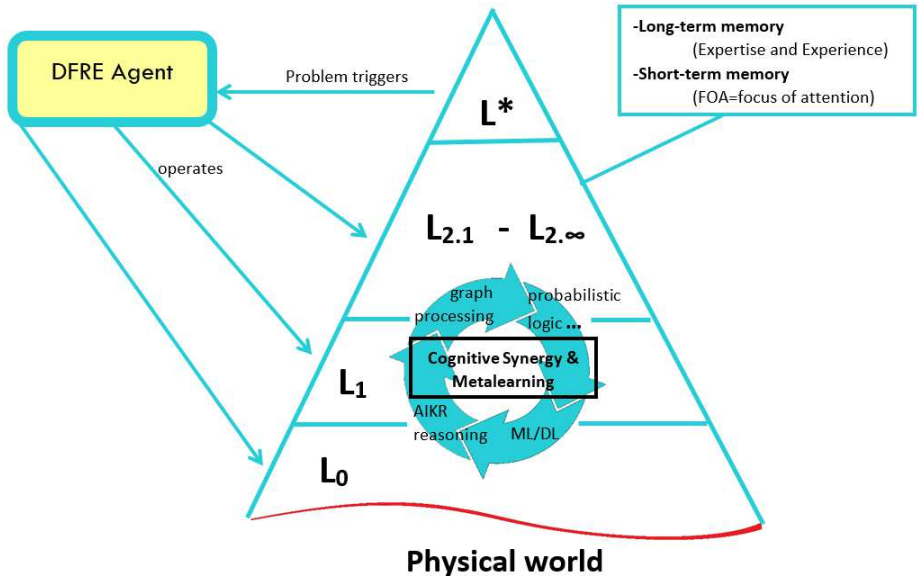}
\caption{\label{fig:loa}DFRE Framework with four levels of abstraction.}
\end{figure}

The DFRE Knowledge Graph (DFRE KG) groups information into four different levels as shown in Figure \ref{fig:loa}. These levels are labeled L0, L1, L2, and L* and represent different levels of abstraction with L0 being closest to the raw data collected from various sensors and external systems and L2 representing the highest levels of abstraction typically obtained via mathematical methods i.e. statistical learning and reasoning.  Each layer can theoretically have infinitely many sub-layers. L* represents the layer where the high-level goals and motivations are stored, such as self-monitoring, self-adjusting and self-repair. There is no global absolute level for a concept and all sub-levels in L2 are relative. However, L0, L1, L2, L* are global concepts themselves. For example, an agent can be instantiated to troubleshoot a problem, such as one related to object recognition or computer networking. The framework promotes cognitive synergy and metalearning, which refer to the use of different computational techniques (e.g., probabilistic programming, Machine Learning/Deep Learning, and such) to enrich its knowledge and address combinatorial explosion issues. 

\begin{figure}[h]
\includegraphics[width=\textwidth]{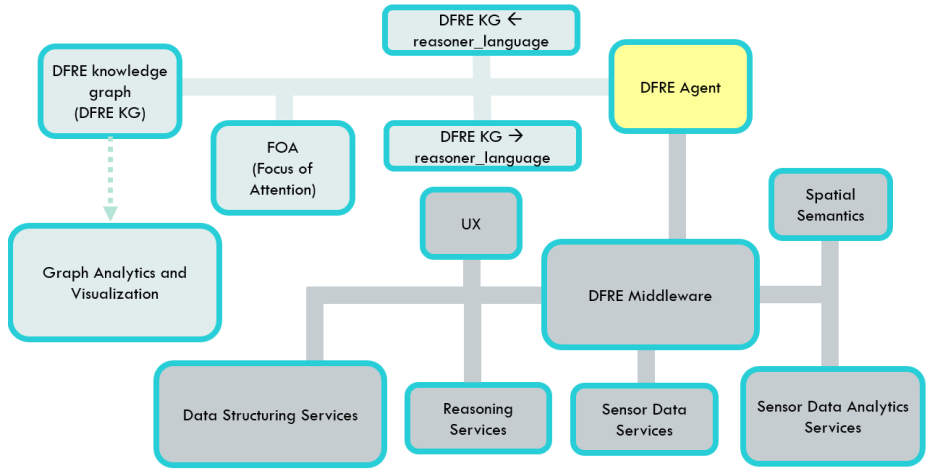}
\caption{\label{fig:architecture}DFRE Framework.}
\end{figure}

One advantage of the DFRE Framework is its integration of human domain expertise, ontologies, prior learnings by the current DFRE KG-based system, and other similar systems, and additional sources of prior knowledge through the middleware services. It provides a set of services that an Agent can utilize as shown in Figure \ref{fig:architecture}. 

The Sensor Data Services are used to digitize any real world data, such as video recordings. Similarly, the Data Structuring Services restructure data if needed, e.g., rectifying an image. These two services are the basis for Image Processing Services which provide a set of supervised and unsupervised algorithms to detect objects, colors, lines, and other visual criteria. The Sensor Data Analytic Services analyze objects and create object boundaries enriched with local properties such as an object's size and coordinates, which create a 2D symbolic representation of the world. Spatial Semantic Services then uses this representation to construct the initial knowledge graph that captures the spatial relations of the object as a relational graph. Any (L2-) high-level reasoning is performed on this knowledge graph. 

Graph-based knowledge representation provides a system with the ability to
\begin{itemize}
    \item Capture the relations in the sub-symbolic world effectively in a world of symbols,
    \item Keep a fluid data structure independent of programming language in which Agents running on different platforms can share and contribute, 
    \item Use algorithms based on the graph neural networks, which allow preservation of topological dependency of information \cite{scarselli2009} on nodes.
\end{itemize}

Graph embedding \cite{cui2018,hamilton2017} is a technique used to represent graph nodes, edges and sub-graphs in vector space that other machine learning algorithms can use. Graph neural networks use graph embeddings to aggregate information from graph structures in non-Euclidean
ways. This allows DFRE Framework to use the embeddings to learn from different data sources that are in the form of graphs, such as Concept Net \cite{speer2017}. Despite its performance across different domains, the graph neural networks suffer from scalability issues \cite{ying2018,zhou2018} because calculating the Laplacian matrix for all nodes in a large network may not be feasible. The levels of abstraction and the focus of attention mechanisms used by an Agent resolve this scalability issue in a systematic way.    

All processes are fully orchestrated by the Agent that catalogues knowledge by strictly preserving the structure while evolving new structures and levels of abstraction in its knowledge graph because, for DFRE KG, knowledge is structure. Multiple agents can have not only individual knowledge graphs but also a single knowledge graph on which  all can cooperate and contribute. In other words, multiple agents can work toward the same goal by sharing the same knowledge graph synchronously or asynchronously. Different agents can have partially or fully different knowledge graphs depending on their experience, and share those entire graphs or their fragments  through the communication channel provided by the DFRE Framework. Note that although the framework can provide supervised machine learning algorithms if needed, the current IoT use case is based on a retail store which requires unsupervised methods as explained in the next section.

\section{Experimental Results}
DFRE Framework was previously tested in the smart city domain \cite{hammer2019}, in which the system learns directly from experience with no initial training required (one-shot) based on a fusion of sub-symbolic (object tracker) and symbolic (ontology and reasoning).The current use case is based on object class recognition in a retail store. Shelf space monitoring, inventory management and alerts for potential stock shortages are crucial tasks for retailers who want to maintain an effective supply chain management. In order to expedite and automate these processes, and reduce both the requisite for human labor and the risk of human error, several machine learning- and deep learning-based techniques have been utilized \cite{baz2016,franco2017,george2014,tonioni2017}. Despite the high success rates, the main problems for such systems are the requirements for a broad training set, including compiling images of the same product with different lighting and from different angles, and retraining when a new product is introduced or an existing product is visually updated.  The current use case does not demonstrate an artificial neural network-based learning. The DFRE Framework has an artificial general intelligence-based approach to these problems.

\begin{figure}[H]
\includegraphics[width=\textwidth]{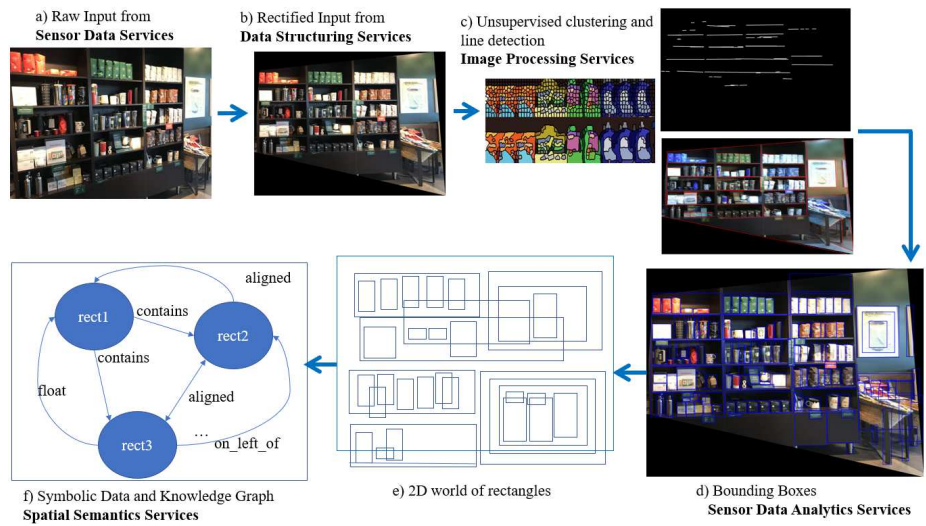}
\caption{\label{fig:retail} Retail use case for DFRE Framework.} 
\end{figure}
Before a reasoning engine operates on symbolic data within the context of DFRE Framework, several services must be run, as shown in Figure \ref{fig:retail}.   The flow starts with a still image captured from a video camera that constantly records the retail shelves by the Sensor Data Services as in Figure \ref{fig:retail}.a, which corresponds to L0 in Figure \ref{fig:loa}. Next, the image is rectified by the Data Structuring Services in  Figure \ref{fig:retail}.b for better line detection by the Image Processing Services, as displayed in  Figure \ref{fig:retail}.c. The Image Processing Services in the retail case are unsupervised algorithms used for color-based pixel clustering and line detection, such as probabilistic Hough transform \cite{kiryati1991}. The Sensor Data Analytics Services in Figure \ref{fig:retail}.d create the bounding boxes which represent the input in a 2D world of rectangles, as shown in Figure \ref{fig:retail}.e. The sole aim of all these services is to provide the DFRE KG with the best symbolic representation of the sub-symbolic world in rectangles. Finally, the Spatial Semantics Services operate on the rectangles to construct a knowledge graph, which preserves not only the symbolic representation of the world, but also the structures within it in terms of relations, as shown in  Figure \ref{fig:retail}.f. This constitutes the L1 level abstraction in the DFRE KG. L1 knowledge graph representation also recognizes and preserves the attributes of each bounding box, such as the top-left \textit{x} and \textit{y} coordinates, the \textit{center}'s coordinates, \textit{height}, \textit{width}, \textit{area} and \textit{circumference}. The relations used for the current use case are \textit{inside}, \textit{aligned}, \textit{contains}, \textit{above}, \textit{below}, \textit{on left of}, \textit{on right of}, \textit{on top of}, \textit{under} and \textit{floating}. Since DFRE KG enforces antisymmetrical relations, the system does not know that \textit{aligned(a,b)} means \textit{aligned(b,a)}, or \textit{on left of} and \textit{on right of} are inverse relations unless such terms are input as expert knowledge or are learned by the system through experience or simulations. As far as the DFRE metamodel, the only innate relations are \textit{distinctions}, which are \textit{anti-symmetric} and \textit{similarity} relations and the rest is learned by experience.

\begin{figure}[H]
    \includegraphics[width=\textwidth]{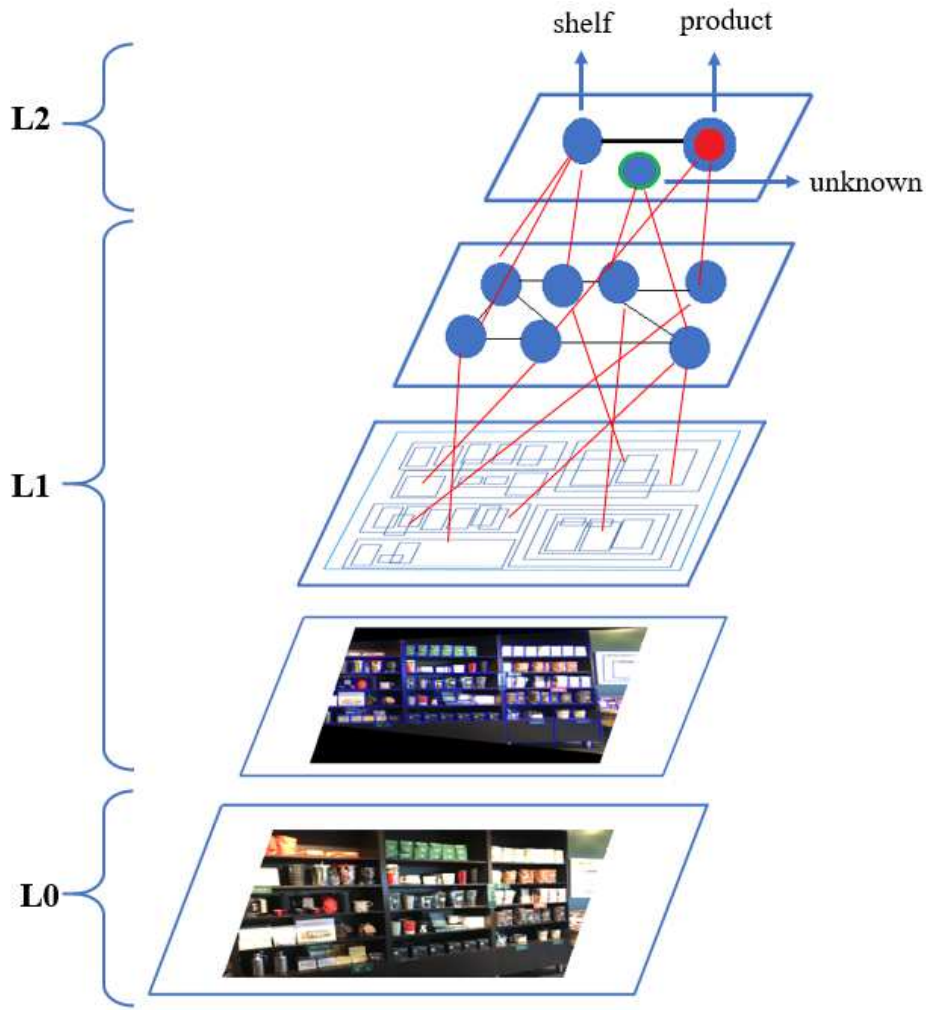}
    \caption{\label{fig:loaretail}LoA for retail use case.}
\end{figure}
The system's ultimate aim is to dynamically determine \textit{shelves}, \textit{products} and \textit{unknown/others} as illustrated in Figure \ref{fig:loaretail} and to monitor the results with timestamps.

While L2 identifies only the concepts of \textit{shelf}, \textit{products} and \textit{unknowns} and the possible relations among them, a reasoning engine, NARS, a non-axiomatic reasoning engine (see \cite{wang2006,wang2010,wang2018} for a detailed review and Narsese-NARS's language), is used to determine their extensions in L1. NARS is a powerful system that operates with the assumption of insufficient knowledge and resources. It also uses an evidence-based truth system in which there is no absolute knowledge.  This is useful in the retail use case scenario because the noise in L0 data causes both overlapping regions and conflicting premises at L1. This noise results from not only the projection of the 3D world input data into a 2D framework, but also the unsupervised algorithms used by L1 services. The system has only four rules for L2 level reasoning:
\begin{itemize}
            \item \textit{If a rectangle contains another rectangle that is not floating, the outer rectangle can be a shelf while the inner one can be a product.} 
            \item \textit{If a rectangle is aligned with a shelf, it can be a shelf too.}
            \item \textit{If a rectangle is aligned with a product horizontally, it can be a product too.}
            \item \textit{If a floating rectangle is stacked on a product, it can be a product too.}
\end{itemize}
Note that applying levels of abstraction gives DFRE Framework the power to perform reasoning based on the expert knowledge in L2 level mostly independent of L1 level knowledge. In other words, the system does not need to be trained for different input and it is unsupervised in that sense. The system has a metalearning objective which is continuously attempting to improve it's knowledge representation. The current use case had 152 rectangles of various shapes and locations, of which 107 were products, while 16 were shelves, and the remaining 29 were other objects. When the knowledge graph in L1 is converted into Narsese, 1,478 lines of premises that represent both the relations and attributes were obtained and sent to the reasoner. Such a large amount of input with the conflicting evidence caused the reasoning engine to perform poorly. Furthermore, the symmetry and transitivity properties associated with the reasoner resulted in scrambling the existing structure in the knowledge graph. Therefore, the DFRE Framework employed a Focus of Attention (FoA) mechanism. The FoA creates overlapping covers of knowledge graphs for the reasoner work on this limited context. Later, the framework combines the results from the covers to finally determine the intensional category. For example, when the FoA utilizes the reasoner on a region, a rectangle can be recognized as a shelf. However, when the same rectangle is processed in another cover, it can be classified as a product. The result with higher frequency and confidence will be the winner. The FoA mechanism is inspired by the human visual attention system, which manages input flow and recollects evidence as needed in case of a conflicting reasoning result. A FoA mechanism can be based on objects attributes, such as color or size, with awareness of proximity. For this use case, the FoA determined the contexts by picking the largest non-empty rectangle and traversing its neighbors based on their sizes in decreasing order. 

DFRE Framework was tested 10 times with and without the FoA. The precision, recall and f-1 scores are exhibited in Table \ref{tab:result}.

\begin{table}[H]
    \centering
    \caption{\label{tab:result}DFRE Framework experimental results.} 
    \begin{tabular}{c|ccc|cll|}
    \cline{2-7}
                                                    & \multicolumn{3}{c|}{\textbf{without FoA}} & \multicolumn{3}{c|}{\textbf{with FoA}} \\ \hline
    \multicolumn{1}{|c|}{\textbf{category}}         & precision     & recall     & f1-score     & precision    & recall    & f1-score    \\ \hline
    \multicolumn{1}{|c|}{\textit{product}}          & 80.70\%   & 29.32\%   & 52.88\%   & 96.36\%   & 99.07\%   & 97.70\%           \\
    \multicolumn{1}{|c|}{\textit{shelf}}            & 8.82\%    & 18.75\%   & 12.00\%   & 82.35\%   & 87.50\%   & 88.85\%           \\
    \multicolumn{1}{|c|}{\textit{other}}            & 36.61\%   & 89.66\%   & 52.00\%   &96.00\%    & 82.76\%  & 88.89\%           \\ \hline
    \multicolumn{1}{|c|}{\textbf{overall accuracy}} & \multicolumn{3}{c|}{\textbf{46.30\%}}           & \multicolumn{3}{c|}{\textbf{94.73\%}}        \\ \hline
    \end{tabular}
\end{table}

The results indicate that the FoA mechanism improves the success of our AGI-based framework significantly by allowing the reasoner to utilize all of its computing resources in a limited but controlled context. The results are accumulated by the framework and the reasoner makes a final decision. This approach not only allows us to perform reasoning on the intension sets of L1 knowledge, which are retrieved through unsupervised methods, but also resolves the combinatorial explosion problem whose threshold depends on the limits of available resources. In addition, one can easily extend this retail use case to include prior knowledge of product types, and other visual objects, tables, chairs, people, shelves, and such, as allowed by the DFRE KG.  

\section{Philosophical Implications}
The nativism versus empiricism debate, which posits that some knowledge is innate and some is learned through experience, was ascribed in the ancient world by the Greek philosophers, including Plato and Epicurus. For the modern world, Descartes is widely accepted as the pioneering philosopher working on the mind as he furthered and reformulated the debate  in the \nth{17} century with new arguments. Perception, memory and reasoning are three fundamental cognitive faculties that enhance this debate by explicating the building blocks of natural intelligence. We perceive the sub-symbolic world, and abstract it in memory, and reason on this symbolic world representation. All three have concept learning and categorization in the center of the human mind. 

The process of concept learning and categorization continues to be an active research topic related to the human mind, since it is essential to natural intelligence \cite{lakoff1984} as well as cognitively inspired robotics research \cite{chella2006,lieto2017}. It is widely accepted that this process is based more on interactional properties and relationships among agents, as well as between an agent and its environment than objective features such as color, shape and size \cite{johnson1987,lakoff1984}. This makes the distinction of anti-symmetric and symmetric relations crucial in DFRE Framework. DFRE Framework assumes that the levels of abstraction are part of innate knowledge. In other words, an agent has L0, L1 and pre-existing L2 by default. This constitutes a common a priori metamodel shared by all DFRE agents. Each agent instantiated from the framework has the abstraction skill based on interactional features and relationships. If the concepts in real life exist in interactional systems, natural intelligence needs to capture these systems of interactions with its own tools, such as abstraction.  These tools should also be based on interactional features by strictly preserving the distinction between symmetry and anti-symmetry. 

The mind is a system as well. Modern cognitive psychologists agree that concepts and their relations in memory function as the fundamental data structures to higher level system operations, such as problem solving, planning, reasoning and language. Concepts are abstractions that have evolved  from a conceptual primitive. An ideal candidate for a conceptual primitive would be something that is a step away from a sensorimotor experience \cite{gardenfors2000}, but is still an abstraction of experience \cite{cohen1997}. A dog fails the mirror test but exhibits intelligence when olfactory skills are needed to complete a task\cite{horowitz2017}. A baby’s mouthing behavior is not only a requisite for developing oral skills but also for discovering the surrounding environment through one of its expert sensorimotor skills related to its survival. The baby is probably abstracting many objects into edible versus inedible higher categories given its insufficient knowledge and resources. What is astonishing about a natural intelligence system is that it does not need a plethora of training input and experiments to learn the abstraction. It quickly and automatically fits new information into an existing abstraction or evolves it into a new one if needed. This is nature’s way of managing combinatorial explosion. Objects and their interconnected relationships within the world can be chaotic. Natural intelligence’s solution to this problem becomes its strength: context. The concept of ‘sand’ has different abstractions depending on whether it is on a beach, on a camera, or on leaves. An agent in these three different contexts must abstract the sand in relation to its interaction with world in its short-term memory. This cumulative set of experiences can later become part of long-term memory, more specifically, episodic memory. DFRE Framework uses a Focus of Attention (FoA) mechanism that provides the context while addressing the combinatorial explosion problem. The DFRE metamodel's  new way of representing practically all knowledge as temporally evolving (i.e. time series) can be viewed as the metamodel's conceptual space. For example, the retail use case given in Section 2 starts with a 3D world of pixels that is abstracted as lines and rectangles in 2D. The framework produces spatial semantics using the rectangles in the 2D world. Based on this situation, a few hundred rectangles produce thousands of semantic relations, which present a combinatorial explosion for most AGI reasoning engines. DFRE KG creates contexts, such candidate shelves, for each scene, runs reasoners for each context, and merges knowledge in an incremental way. This not only addresses the combinatorial explosion issue, but also increases the success rate of reasoning, provided that the levels of abstraction are computed properly \cite{gorban2018}.

Abstracting concepts in relation to their contexts also allows a natural intelligence to perform mental experiments which is a crucial part of planning and problem solving. DFRE Framework can integrate with various simulators, re-run a previous example together with its context, and can alter what is known for the purposes of  experimentation to learn new knowledge, which is relationships and interactions of concepts. 

Several mathematical models and formal semantics \cite{duntsch2002,belohlavek2004,wang2008,wille1982,ma2007} are proposed to specify the meanings of real world objects as concept structures and lattices. However, they are computationally expensive \cite{jinhai2015}. One way to overcome this issue is with granular computing \cite{yao2012}. The extension of a concept can be considered a granule and the intension of the concept is the description of the granule. Assuming that concepts have granular common parts with varying derivational and compositional stages and categorization, abstraction and approximation occur at multiple levels of granularity which has an important role in human perception \cite{hobbs1985,yao2001,yao2009}.

Having granular structures provides structured thinking, structured problem solving, and structured information processing \cite{yao2012}. DFRE Framework has granular structures but emphasizes the preservation of structures in knowledge. Being in the extension of a concept does not necessarily give the granular concept the right to have similar relations and interactions of its intensional concept up to certain degree or probability. Each level must preserve its inter-concept relationships, and its symmetry or anti-symmetry. When a genuine problem that cannot be solved by the current knowledge arises, it requires scrambling the structures and running simulations on the new structures in order to provide an agent with creativity. Note that this knowledge scrambling is performed in a separate sandbox and the DFRE Framework ensures that the primary DFRE KG is not corrupted by these creative, synthetic, ``knowledge-scrambling" activities.

\section{Conclusion}
We have outlined key aspects of the DFRE Framework based on a simple idea that knowledge is hierarchical structure. While this paper provides highlights of one experiment in the visual domain employing an unsupervised approach, we have also run similar experiments on time series and natural language data with similar promising results. Based on the level of approvals we are able to obtain, our plan is to release more details including mathematical formulations of the DFRE Framework, source code, and other artifacts. At a minimum, prior versions of the DFRE Framework utilized in the smartcity domain should soon be released as open source. 

We are always looking for collaborators (academic or commercial). Feel free to contact the authors with your comments and suggestions.

\end{document}